**Robust Localization of an Arbitrary Distribution of Radioactive Sources for Aerial Inspection – 18094**

Dhruv Shah [1], Sebastian Scherer [2]
[1] Indian Institute of Technology, Bombay
[2] Carnegie Mellon University

**ABSTRACT**

Radiation source detection has seen various applications in the past decade, ranging from the detection of dirty bombs in public places to scanning critical nuclear facilities for leakage or flaws, and in the autonomous inspection of nuclear sites.

Despite the success in detecting single point sources, or a small number of spatially separated point sources, most of the existing algorithms fail to localize sources in complex scenarios with a large number of point sources or non-trivial distributions & bulk sources. Even in simpler environments, most existing algorithms are not scalable to larger regions and/or higher dimensional spaces. For effective autonomous inspection, we not only need to estimate the positions of the sources, but also the number, distribution and intensities of each of them.

In this paper, we present a novel algorithm for the robust localization of an arbitrary distribution of radiation sources using multi-layer sequential Monte Carlo methods coupled with suitable clustering algorithms. We achieve near-perfect accuracy, in terms of F1-scores ($> 0.95$), while allowing the algorithm to scale, both to large regions in space, and to higher dimensional spaces (5 tested).

**INTRODUCTION**

With the large number of aging nuclear sites and the need for the proper management and disposal of radioactive waste, there is an increasing demand for robust mechanisms for the autonomous inspection of the disposal sites to analyze the waste. Such a mechanism requires the *multi-modal mapping* of the environment, including an accurate estimate of the distribution and intensities of the radioactive materials, using either ground or aerial robots. To enable autonomous inspection, we would like the robot to identify regions of interest, based on visual imagery, thermal imagery, radiation strengths, waste estimates etc., and map such regions with greater precision.

To generate a radiation map, the robot needs to identify the position and estimates of the sources of radioactivity. Most of the existing algorithms only localize multiple point sources, and don't extend to a large number of such sources [1, 2, 3]. The problem of localizing bulk sources and complex distributions is not addressed in literature.

The availability of gamma-imaging cameras simplifies localization, by generating a likelihood map of the source estimates. These sensors, however, (i) require very large exposure time, during which the robot must remain stationary, and (ii) are too bulky for a flying robot (~2-4 kg). These two factors eliminate the possibility of mounting such a sensor on light-weight drones or quadcopters, and hence we are required to move to the simpler solid-state detectors, which require lower exposure time (few seconds), and are light-weight (~50grams).

Our work focuses on developing a scalable algorithm for the robust localization of multiple radiation sources, irrespective of the distribution, using cheap light-weight *particle flux* detectors. We also look at various extensions to the simple source localization problem, by attempting to solve more complex scenarios, without changing the core of the algorithm. We propose an improved version of a hybrid formulation of a particle filter (sequential Monte Carlo method) and clustering techniques to address and tackle the challenges. In our method, we begin by generating hypotheses about source positions (particles); when a new measurement is received, the likelihoods of the hypothesis to be a real source is evaluated. Unlike a traditional particle filter, we *selectively* evaluate the particles based on their influence at the sensor that generated the reading. We also allow for multiple source localization by updating the *prediction* stage of the particle filter, and by reinforcing the algorithm using partially resolved sources in the environment. This notion of partially resolving the sources is key to the performance of the algorithm in complex scenarios. Our main contributions are as follows:

- We propose an improved localization approach for multiple sources in complex environments and distributions that (i) can resolve large number of sources efficiently, (ii) is scalable to large regions in space and to higher dimensions, and (iii) can be used to identify and localize bulk radiation sources.
- We introduce the notion of *candidate sources*, which are source parameter estimates obtained after clustering. These are evaluated based on a confidence metric, before labeling. This allows the algorithm to have a very high accuracy.
- The algorithm allows for partial localization of the sources in the environment, which is used to reinforce further iterations, accelerating the process. In this way, the problem of localization of large, complex environments is sequentially broken down to smaller simpler scenes, allowing improvement in performance over time.

The balance of the paper is organized as follows. We begin by talking about the existing work in the field of radiation source localization. Next, we formulate the problem mathematically and introduce the concept of Bayesian filtering & recursive Bayesian methods before presenting the proposed algorithm in detail. The following section presents results of some of the experiments performed, by extending it to higher dimensional spaces and real-world environments. This is followed by the quantitative analysis of the algorithm and a comparative study against the state-of-the-art.

**RELATED WORK**

The problem of $1/R^2$-type source localization has been analyzed quite extensively in literature, in the contexts of acoustics, radio transceivers, electromagnetic fields, chemical plumes and also for radiation sources. The classical problem of single source localization with known source intensity has been solved by least squares fitting [4] and maximum likelihood estimation (MLE) methods, which search the parameter space for the most likely source parameters [5]. Authors in [6] adapt a time difference of arrival (TDoA)-based algorithm in log-space to exploit the logarithmic differences in source strength measurements to infer the source position. These algorithms assume prior knowledge on the source intensity, and hence, are impractical in a mapping setup. Baidoo-Williams et al. [1] use a ML-localization framework for the localization of a source with unknown intensity, guaranteeing no false stationary points as long as the source lies in the open convex hull formed by the sensors. These algorithms are not applicable in scenarios involving multiple point sources.

For localizing multiple sources, most existing algorithms estimate the number of sources $N_S$, in addition to the existing source parameters. Morelande *et al.* [7, 2] begin by estimating the number $N_S$ using Gaussian mixture model based selection, and then compute the source parameters using the MLE method. As quoted in [2], the accuracy of the model selection degrades with increasing $N_S$; also, the runtime explodes with the number of sources. A similar approach is used in [8] ,where targets are modeled with Gaussian mixture models, followed by Akaike's/Bayesian Information Criteria to estimate $N_S$. This is followed by a simple expectation maximization (EM) routine and clustering, to estimate the parameters. As quoted in [2], these EM or MLE based algorithms do not scale beyond four sources.

In [3], the authors propose the multiple source localization by performing the gradient descent optimization of non-convex cost functions, assuming that the sources lie in the convex hull formed by the sensor positions. In [9], the authors propose to solve the localization problem using convex optimization, assuming that the sources are located in a grid over the region of interest – the algorithm proceeds by discretizing the search space and localizing in the discretized locations. In a reported case with one source and 196 sensors, the algorithm takes *209s* to converge, which explodes to *3205s* with 4 sources – indicating that the algorithm fails in scaling to more complex distributions.

Another interesting approach was proposed [10] using partial Bayes factors to evaluate the models. The use of Monte Carlo methods allows for powerful implementations using traditional particle filters [11], which are used to approximate the source distribution for complex environments. Although this implementation does not scale very well, it served as a good alternative for the source parameter estimation. With updates to the estimation stage above, *Rao et al.* [12] propose a particle filter that can be used for localizing a small number of point sources efficiently, and can scale to moderately large regions.

**SETUP**

This section details the setup of the problem and leads towards our sequential Monte Carlo implementation. For the sake of illustration, we assume the environment to be a two-dimensional grid containing point sources, which shall later be extended to higher dimensional spaces. We start off by formulating the problem and the sensor model used, and then give a brief on recursive Bayesian estimation before proceeding towards the algorithm.

**Problem Formulation**

We consider the localization of $N_S$ radiation point sources of unknown strengths using a mobile ground/aerial robot in a two-dimensional plane around the target area with obstacles. Continuing the notation scheme used by [12], let $\mathcal{A} = \{A_1, A_2, ..., A_{N_s}\}$ denote the set of radiation sources. Each source is parametrized by a three-value vector $A_k = <A_k^x, A_k^y, A_k^{str}>$, for $1 \leq k \leq N_s$. The parameters, thus, define the position $(x, y)$ in cm and strength $A^{str}$ in microCuries ($\mu Ci$) of the radiation source in concern. In the general case, the parameter vector $A_k$ can be extended to account for position in 3D ($A_k^z$) or for more complex source distributions by appending terms corresponding to dipole and quadrupole moments of the source (described later).

The task of measurement of radiation intensity can be classified as the task of measuring either of the following: particle flux, energy fluence, beam energy, Kerma or dose. Each of these present different approaches to representing the ionization strength of the emitter. For this research, we suppose the simplest model for measurement - measuring the *particle flux* or source influence at different points in space.

In a surveillance area without obstacles, the contribution of $A$ in the intensity recorded at a location $x$ is

$$\mathcal{P}(x, A) = A^{str} \ (h^2 + |x - A^{pos}|^2)^{-1} \tag{1}$$

where $h$ is the height of the sensor from the ground, and $A^{pos} = (A^x, A^y)$. The values $(h, A^x, A^y)$ are obtained from the odometry measurements in the reference frame of the robot. Eqn. (1) is a model widely used in existing work and has been verified experimentally [2, 13].

**Sensor Modeling**

The most common type of sensor is the *Geiger counter*, which identifies alpha, beta and gamma rays using the ionization effect in produced in a Geiger-Muller tube. Scintillators, as the name suggests, are excited by ionizing radiation, absorb its energy and scintillate. This class of devices measures the particle flux, which is basically a measure of the strength of the radiation source at the point of measurement.

Another class of radiation sensors comprise portable gamma imaging cameras. One such instrument is the *Polaris-H*, which uses layers of *CdZnTe* crystals to give a cone of likelihood of the source [14], instead of simply giving the flux measurement. Such sensors tend to be bulky (~2-4kg) and require exposure times of the order of 10-20 minutes to compute accurate imagery. Due to these limitations, we restrict ourselves to using point radiation sensors, measuring particle flux, for the rest of this report. In particular, solid state detectors and scintillators can be modeled as shown below.

Instead of a sensor network, as used in standard distributed approaches [12, 15, 16], a sensor/sensor-array is mounted on the mobile robot, which traces a trajectory of choice. The intensity is usually reported in counts per second (CPS) or counts per minute (CPM). Let $p_i$ denote the position at which a sensor reading was taken, in the ground frame. If the robot has multiple sensors, it may record multiple readings from the same position of the robot, but the sensors would be at distinct positions in the ground frame. The sensor, at each position $i$, will record a background radiation $B_i$ that is naturally present due to cosmic rays and naturally occurring radio isotopes. This reading is usually very low (few CPM) and should not affect the model, but has been considered to demonstrate robustness. The sensors can have different efficiencies in counting the number of ionizations (hereby referred to as *interactions*) due to the difference in manufacturing technologies, processes, sizes etc. This can be incorporated using a calibration constant $E_i$. Given all of the above, the expected intensity (in CPS) at $p_i$ can be given by

$$I_i = 3.7 \times 10^4 \times E_i \sum_{j=1}^{N_s} \mathcal{P}(p_i, A_j) + B_i \tag{2}$$

The constant $3.7 \times 10^4$ is the conversion factor from microCurie to CPS (Becquerel). Given the expected intensity $I_i$, the measurements received by the sensor $i$, $m(S_i)$ are modeled as a Poisson process with average rate $\lambda = I_i$, which is known to be the distribution followed by photon intensities [17].

**Bayesian Estimation**

The problem of source localization can be viewed as that of the mapping problem in simultaneous localization and mapping (SLAM). Hence, we model the scenario in a similar fashion. Consider the general state space model with hidden variables $\boldsymbol{h_t} = \{h_0, h_1, \dots, h_t\}$ and observed variables $\boldsymbol{o_t} = \{o_1, o_2, \dots, o_t\}$; we would like to perform an inference on the hidden variables $\boldsymbol{h_t}$. Given the observed variables, Bayesian inference on the hidden variables relies on the joint posterior distribution $p(\boldsymbol{h_t}|\boldsymbol{o_t})$. Assuming that the hidden variables have some initial distribution $p_0(h_0)$, a transition model $p(h_t|h_{t-1})$, and that the observations are conditionally independent given the hidden process (yielding the marginal distribution $p(o_t|h_t)$), a Bayesian filter can be used to derive a recursive expression for this posterior.

$$p(\boldsymbol{h_t}|\boldsymbol{o_t}) = p(\boldsymbol{h_{t-1}}|\boldsymbol{o_{t-1}}) \frac{p(\boldsymbol{o_t}|\boldsymbol{h_t})p(\boldsymbol{h_t}|\boldsymbol{h_{t-1}})}{p(\boldsymbol{o_t}|\boldsymbol{o_{t-1}})} \tag{3}$$

A formal proof to Eqn. (3) can be found in [18].

The above can be extended to a SLAM framework with the robot position and map as hidden variables $h_t = \{x_t, \Theta\}$, and the sensor measurements $z_t$ as a recursive update:

$$p(x_t, \Theta \mid z_t) = \eta \; p(z_t \mid x_t, \Theta) \int p(x_t \mid x_{t-1}) \, p(x_{t-1}, \Theta|z_{t-1}) dx_{t-1} \tag{4}$$

which indeed has the form of the recursive Bayesian filter, Eqn. (3).

Unfortunately, the computation of the integral over all robot positions $x_{t-1}$ is intractable and unfeasible, and hence, must be approximated. The most popular approximations involve the use of Kalman filters or particle filters. *Particle filters* form a sequential Monte Carlo approximation to the recursive Bayesian filter described above. In addition, particle filters provide a usable implementation of Bayesian filtering for systems whose belief state and sensor, process noise can be modeled by non-parametric probability density functions [19].

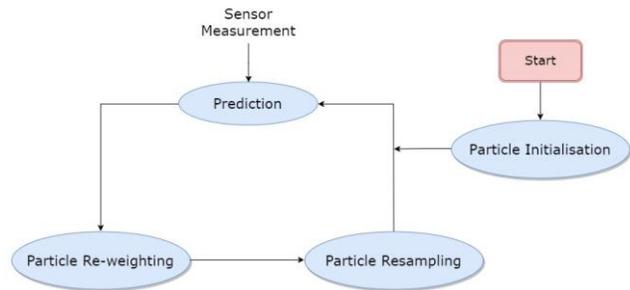

**Figure 1**: Algorithmic flow for a traditional particle filter

In the traditional case, the particle filter can maintain a discrete approximation of the SLAM posterior using a large set of *particles*, or samples, in the state space. In this case, a weighted approximation of the recursive Bayesian filter can be given as:

$$p(s_t|z_t, u_t) \approx \sum_{i=1}^{N_s} w^{(i)} s_t^{(i)} \quad (5)$$

where we have a set of $N_s$ particles $\{s^{(1)}, s^{(2)}, \ldots, s^{(N_s)}\}$, with weights $w^{(i)}$. Assuming that we have such a set of weights and particles, we can update these particles by drawing samples from $q(.)$:

$$s_t^{(i)} \sim q(s_t|s_{t-1}^{(i)}) \quad (6)$$

where $q(x_t|x_{t-1}, z_t)$ is a distribution that is easy to sample, and approximates $p(x_t|x_{t-1}, z_t)$ with fatter tails, assuring the coverage of $p(.)$. Thus, the weights are updated as:

$$w_t \propto w_{t-1} \frac{p(z_t|s_t) \, p(s_t|s_{t-1})}{q(s_t|s_{t-1}, z_t)} \quad (7)$$

In the optimal case, $q(.) = p(.)$, but since it may not be easy to sample from that distribution, we instead use the motion model $p(x_t|x_{t-1})$, which allows us the to simplify the weight update equation to

$$w_t \propto w_{t-1} \, p(z_t|s_t) \quad (8)$$

It is key to note that the core prediction step of the Bayesian filter (Eqn. 5) poses a fundamental limitation in localizing multiple sources, which was modified in [12]. We build upon this improved model of the traditional particle filter (Fig. 1) in the next section.

**METHODS**

In this section, we describe our algorithm for the robust localization of an arbitrary distribution of radiation sources using particle flux measurements from a mobile robot. The proposed algorithm recursively refines the source parameter estimates based on newly acquired measurements, and can localize fairly complex source distributions recursively. Fig. 2 outlines the flow of the algorithm. The algorithm begins by spawning a collection of particles, randomly or based on a known prior, each of which hypothesizes the location and strength of a radiation source. At each new robot position, the algorithm identifies the particles in its *area of influence* and their weights are updated in a Bayesian manner, according to the newly received sensor measurement, and the prior weights. After weighting the particles, a resampling procedure normalizes the weights of the particles. This procedure is repeated as a new measurement arrives. At any point in this cycle, a list of candidate sources can be obtained by running a suitable labeling routine, which estimates the most likely source parameters.

The algorithm can reinforce its estimates after some of the sources have been resolved, enabling better accuracy for localizing a large number of sources. The outer loop, as seen in Fig. 2 illustrates how the algorithm improves with time, eventually localizing the whole scene. Some salient features of our

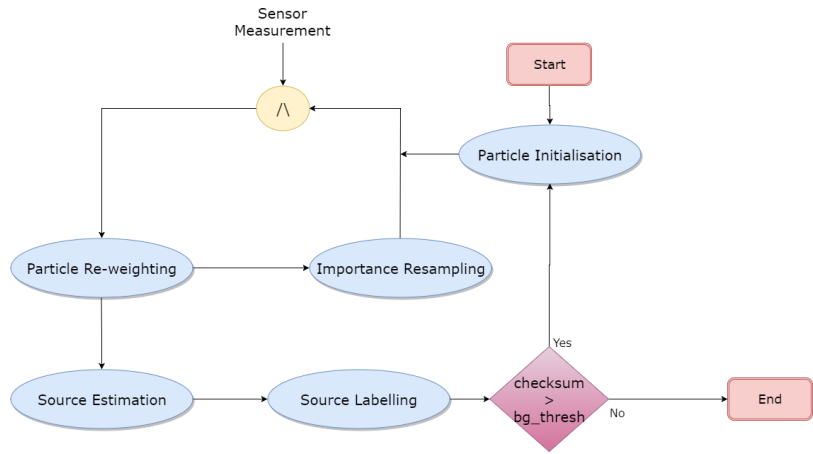

**Figure 2:** Overview of the proposed algorithm

formulation are as follows:
- We use the notion of a *time step*, which refers to the unit of time when the readings from all positions have been received and processed, i.e., the robot has successfully explored the environment.
- Instead of explicitly modeling different source distributions and interactions, we compute the parameter estimates for the particles independent of each other. This enables the algorithm to scale readily with the number of sources, unlike the exponential trends demonstrated by conventional methods [19, 5].
- At any point in time, the source parameters can be computed by a suitable clustering algorithm. A candidate source with confidence score above a threshold is declared as *resolved*, and the algorithm uses that information in the subsequent steps. This allows more efficient localization of multiple sources, and at the same time, enables the algorithm to detect complex source distributions that would not be detected otherwise.
- At no point in time, does the algorithm require any a priori information on the distribution. However, if such information is available for the source strengths, locations, or number, the convergence can be sped up by a great amount. The algorithm, despite being independent of any such assumptions, can be readily refined on provision of more information.
- Without loss of generality, the parameter vector of a particle can be arbitrarily high-dimensional, including the z-coordinate, a dipole moment and dipole moment vector and so on. In principle, the algorithm would converge to the correct result in each case, but the number of time steps involved & particle population required would depend directly on the number of source parameters to be estimated.

**Particle Initialization**

At time step $t = 0$, the particle filter is initialized as follows. Let $\mathcal{P} = \{\boldsymbol{p}_i^{(t_i)} | i = 1, \ldots, N_\mathcal{P}\}$ be a set of particles in the target area. Each particle $p_i^{(t_i)}$ is a vector in the parameter space $\mathcal{A}$, denoting the position and source parameters of the hypothesized source. For the simplest case, it can be visualized as the three-value vector in $\mathcal{A}$. The superscript $t_i$ is an integer denoting the time step at which the particle is being updated. In the general formulation, with no assumption on source strengths and positions, we initialize $\mathcal{P}$ with uniformly random particles in the target area and a large window of prospective source strengths. If prior knowledge is available, the particles can be initialized according to the pre-existing distribution, greatly improving the performance of the algorithm. Some examples of guided initialization are shown in Fig. 3.

We denote the cardinality of $\mathcal{P}$, or the number of particles as $N_\mathcal{P} = |\mathcal{P}|$. $N_\mathcal{P}$ directly governs the coverage of $\mathcal{P}$ in $\mathcal{A}$, in a random initialization. A larger coverage will result in a more accurate estimate in a shorter time, because the Monte Carlo methods approximate the real distributions only when the number of particles being sampled is sufficiently large. In addition to the parameter vector, we associate each particle $p \in \mathcal{P}$ with a weight $w(\boldsymbol{p})$ such that $w(\boldsymbol{p}) \approx \sum_j P(A_j = \boldsymbol{p})$ and $\sum_{p_i \in \mathcal{P}} w(\boldsymbol{p}_i) = 1$. This weight measures the probability that an actual radiation source can have the same parameters as the concerned particle. Since the number $\sum_j P(A_j = \boldsymbol{p})$ cannot be computed, it must be approximated. In the next subsection, we shall see how the weights are updated, so as to converge to the required number. For the initialization stage, each particle is assigned equal weight, given by $w(\boldsymbol{p}) = \frac{1}{N_{\mathcal{P}}} \; \forall \; \boldsymbol{p} \in \mathcal{P}$.

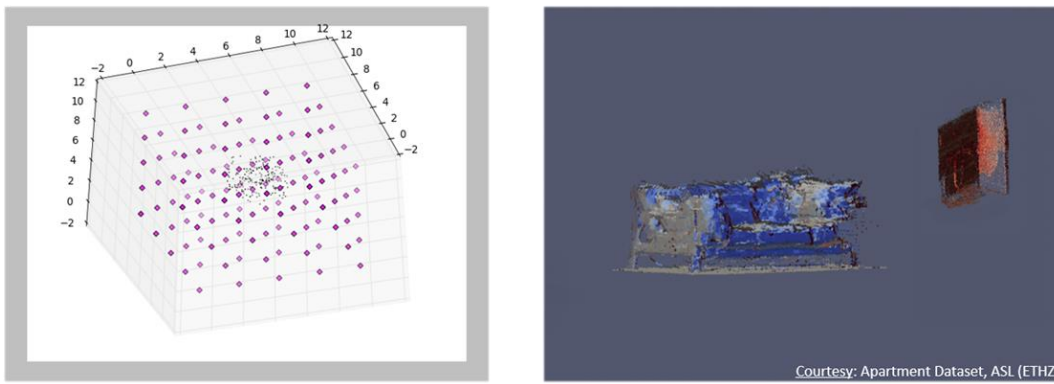

**Figure 3**: Examples of guided particle initialization (a) in a simulated 3-D environment, and (b) using point cloud data

**Particle Reweighting**

This step forms the Bayesian estimation framework of the filter. Although we do not know the actual source parameters, we can estimate this probability using the sensor measurements. The particle filter under consideration is a sequential Monte Carlo approximation to the recursive Bayesian filter, as formulated in the Setup section. Rephrasing Eqn. 8, the weight of a particle $\boldsymbol{p}_i$ can be updated as

$$w(\boldsymbol{p}_i^t) = p^*(m(S_i)|\boldsymbol{p}_i^t) \cdot w(\boldsymbol{p}_i^{t'}) \tag{9}$$

Here, the distribution $p^*(.)$ represents the normalized version of $p(.)$, the distribution obeyed by the radiation source. In particular, $p(x|y)$ stands for the probability of obtaining a value $x$, when sampling from a Poisson distribution with mean $y$. For this case, $p^*(x|y)$ can be given as

$$p^*(x|y) = \frac{p(x|y)}{f_y} \tag{10}$$

Here, $f_y = p(\lfloor y \rfloor | y)$ is the normalizing factor and $\lfloor . \rfloor$ depicts the floor operator. If we have prior information on any known sources, or some sources $(A_k^*)$ have been resolved by the algorithm, Eqn. 9 can be modified to incorporate this information, and hence converge more efficiently.

$$w(\boldsymbol{p}_i^t) = p^*(m(S_i)|\boldsymbol{p}_i^t, (A_k^*)) \cdot w(\boldsymbol{p}_i^{t'}) \tag{11}$$

Updating the weights of the particles, as above, for all particles in the target area, for each sensor, would be both computationally infeasible and impractical. For particles far away from a sensor, a small amount of uncertainty due to the sensor would cause a large reduction in weight, leading to the loss of a potential candidate. This is where the notion of *fusion range*, first proposed by Rao *et al*. [12] comes handy. The notion of a fusion range is directly linked to the idea of area of influence of a source or sensor. The influence of a source at a given point falls off with the second order of the distance between them. This means that sensors sufficiently far enough do not exhibit a strong enough influence to have a say in the weight of a particle. Fig. 4 provides a simple illustration of the phenomenon. Henceforth, we define the fusion range as a number $d$ such that $\mathcal{P}' = \{ \boldsymbol{p}_i \mid \| S_i - p_i^{pos} \|^2 \leq d^2 \}$ is the set of particles lying within the fusion range of a sensor at location $S_i$. Thus, Eqn. 9 is used to update the weights of every particle $\boldsymbol{p} \in \mathcal{P}'$. The sensor measurements close to the particles will provide a better view at those locations. By only updating particles that are close to the sensor concerned, we allow multiple sources to co-exist in a small neighborhood. Along with improving the accuracy of the estimation, this also enables us to increase the computational efficiency of the algorithm.

In addition to the above, we maintain an entry called *checksum* for each time step – which stores the sum of readings registered by each sensor position, and can be seen as a simple method to identify the end point of source localization - if the checksum is negative (or *very small*), the algorithm has localized all sources.

**Importance Resampling**

Importance resampling, or sampling importance resampling (SIS), are common practices used with Monte Carlo methods for the resampling stage of the filter [20]. In this step, we wish to replace particles of low weights with particles of higher weights, as a way to ensure a better convergence and prevent *particle degeneration*. In a particle filter without resampling, all the particles will have decreasing weights over time, except for the one closest to the source(s). Not only will this give an inaccurate estimate of the source parameters, it can also give spurious results.

Since the weight update was carried out only on particles within the fusion range $\mathcal{P}'$, the resampling shall also be done on the same set. The weights of the particles, defined and updated as per earlier sections, act as the importance distributions for the intended resampling procedure. This resampling is accomplished by sampling from $\mathcal{P}'$ with probabilities $\frac{w(\boldsymbol{p}_i)}{\sum_\mathcal{P} w(\boldsymbol{p}_j)}$ for all $p_i \in \mathcal{P}'$. The resampled particles are then assigned uniform weights. This new set is merged with the original set of particles to give the updated set of particles.

During the resampling process, we introduce zero-mean Gaussian noise into each parameter of the generated particle. This noise serves a two-fold purpose: (i) it allows propagation of the particle in parameter space, allowing convergence to points that were not initialized in $\mathcal{A}$, and (ii) it ensures that particles are always slightly different, so that the filter does not always degenerate to a single point. The Gaussian nature of the noise can be exploited in the labeling stage, where a clustering technique like the mean-shift algorithm, which uses a multi-variate Gaussian kernel for grouping, can be fine-tuned to obtain robust classification.

The above resampling procedure eliminates particles that do not correspond to any real sources, because their weights degrade over time, and hence vanish. As time proceeds, areas with no radiation sources in the vicinity would have little or no particles around them. If the environment was dynamic, and a new source was to move into that region, it could go undetected. To account for this scenario, we randomly replace a small percentage of particles, say 5-10%, with random particles. This ensures that the new sources also have a chance at being detected.

**Source Estimation**

Given the set of particles and associated weights at any given point in time, we can compute estimates of source parameters by running a suitable clustering and evaluation technique. We run the source estimation stage after every $s$ time steps of running the inner loop, unlike the traditional particle filter or the filter proposed in [12], which runs the prediction phase at each sensor reading. This not only saves on computation time, but also ensures that the estimates are computed only after a significant portion of the map has been explored with high confidence.

We generate source parameter estimates from the particles by clustering them, and then representing the whole set by their respective cluster centroids. The clustering of the particles in $\mathcal{A}$ can be done by any of the following methods:

- **Mean-shift Clustering**: This is a non-parametric feature-space algorithm widely used to locate the peaks of a density function [21]. It proceeds by fitting multi-variate Gaussians over the feature space and labeling the lobes as clusters. This step allows for multiple clusters, and hence enables the algorithm to localize multiple radiation sources. In addition, since the mean-shift technique groups based on the kernel parameters, and not the number of clusters, no information on the number of sources is required. A particular reason why this method works quite accurately is its robustness to Gaussian noise; since the kernel is a multi-variate Gaussian, the clustering is robust to Gaussian noise due to the environment and measurement.
- **Hierarchical Clustering:** Agglomerative hierarchical clustering (AHC) can be used for the classification of particles in the Euclidean space. AHC begins by placing each particle in a cluster of its own, and groups them into larger clusters, based on proximity in $\mathcal{A}$. This can be particularly useful for large particle populations, when mean-shift becomes computationally unfeasible. Various

- algorithmic manipulations [22] allow agglomerative AHC to run in a worst case of $\mathcal{O}(n^2)$ which makes up for the sub-optimal clustering.
- **ID-based Clustering:** Another naive way of grouping particles can be based on the particle ID, a unique number given to every particle that was originally initialized. When resampling occurs, particles are respawned and may undergo a change in ID. Grouping based on this ID can serve as an inexpensive way to identify different clusters. This works amazingly well for very large environments, with a sparse distribution of sources - a case where the large particle population would make both the above algorithms computationally expensive. For denser environments, this technique fares poorly

We propose the notion of *candidate sources*, which is the term used to refer to each of the cluster centroids identified by the clustering technique. These candidate sources are then screened, to eliminate spurious estimates and help improve the accuracy of classification. This also gives us the liberty to maintain a generic clustering algorithm that results in a large number of hypotheses, which can then be evaluated & classified autonomously by the algorithm.

**Source Labeling**

The last step of our algorithm evaluates the quality of candidate sources suggested by the clustering stage and decides whether the candidate can be labelled as a source, or dropped. We define the k$^{th}$-order confidence metric $c_k$ for a candidate source $\boldsymbol{p}_c$ as follows.

$$c_k = \sum_{i=1}^{k} \omega_i \, p^*(\, m(S_i) \mid I(S_i, \boldsymbol{p}_c)) \tag{12}$$

where $S_i$, $1 \leq i \leq k$ are the $k$ sensors nearest to the candidate source, $p^*(.)$, $I(.)$ are same as defined earlier in Eqn. 10 and Eqn. 1, and $\sum_i \omega_i = 1$. The weights, and $k$, are manually chosen, and can be optimized for best results.

This confidence measure is employed to test the credibility of the candidate source and decide on its acceptability – a candidate is declared as a resolved source *iff* it's confidence score exceeds a predefined threshold. The knowledge of these resolved sources is then incorporated into the core algorithm as per Eqn. 11, to reinforce the computation and accelerate the localization of the unresolved sources.

**The Complete Algorithm**

We propose an algorithm that can localize multiple radiation sources sequentially. After the source estimation has been performed, candidate sources are evaluated based on the confidence metric $c_k$ and labeled accordingly. For closing the loop, and provide an end point to the algorithm, when necessary, we follow the simple routine given by Alg. 1.

Here, *bg_thresh* refers to the background radiation count of the environment and *checksum* is used as defined earlier. The labelled sources are now classified as known sources and the algorithm uses them to localize sources further, as suggested in Eqn. 11.

**Algorithm 1:** Closing the Loop

```
1  Function label_sources()
2      for p_c in candidate_sources do
3          if p_c^s ≥ source_thresh then
4              obtain confidence measure c_k from Eqn. (13);
5              if c_k ≥ confidence_thresh then
6                  resolved_sources ← p_c;
7      if checksum ≥ bg_thresh then
8          rerun the entire algorithm;
9      else
10         Exit algorithm (end point)
```

**RESULTS**

In this section, we present the results obtained by extending the algorithm described earlier to a variety of cases. We begin by demonstrating a sample run of the algorithm, as listed in the previous section, on simple cases.

Fig. 5 illustrates a simple scenario with multiple sources. Consider a $10m \times 10m$ environment, with 3 point sources. Measurements are taken from a ground robot exploring the room in a lawnmower pattern. We proceed with random initialization, and exit the inner loop after 3 time steps. The estimation stage identifies 7 candidate sources (narrowband kernel chosen), which are then evaluated to resolve all the sources in a single iteration.
*(Confidence Score $c_3 = \{96\%, 90\%, 98\%\}$; Runtime $= 33.2s$)*

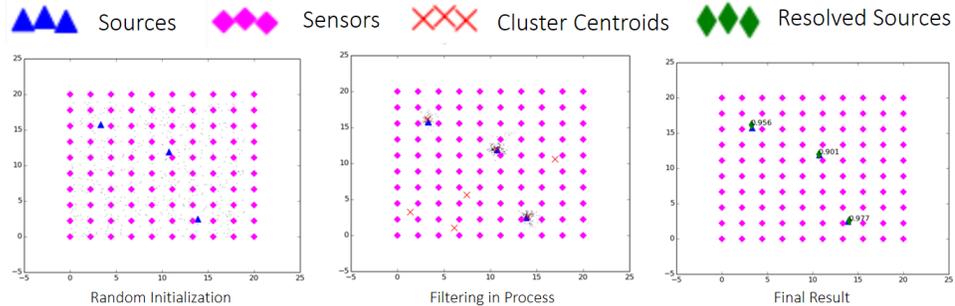

**Figure 5:** Sample run 1 of the algorithm in 2-D space, with 3 sources and 100 readings
(*Algorithm terminates in 3 time steps*)

Fig. 6 demonstrates a scenario in which the proposed algorithm fares well in localizing multiple sources. The setup is like the previous example, with 3 sources randomly positioned in the same grid. After 3 time steps, the algorithm identifies 8 clusters, and on evaluating them using $c_3$, the algorithm only manages to localize 2 out of these; the third source was not captured with a good confidence score. Hence, the two sources are marked as resolved, and the inner loop is called again. After a second round of filtering, the algorithm manages to localize the third source with much higher confidence of 93.7%, establishing our idea. After a portion of the sources have been localized, the localization task is, in fact, simpler, and thus the efficiency of the algorithm improves. State-of-the-art implementations [12, 2, 7] fail to capture the third source, or do so with a much higher localization error. *(Confidence Scores $c_3 = \{87\%, 85\%, 94\%\}$; Runtime $= 64.0s$)*

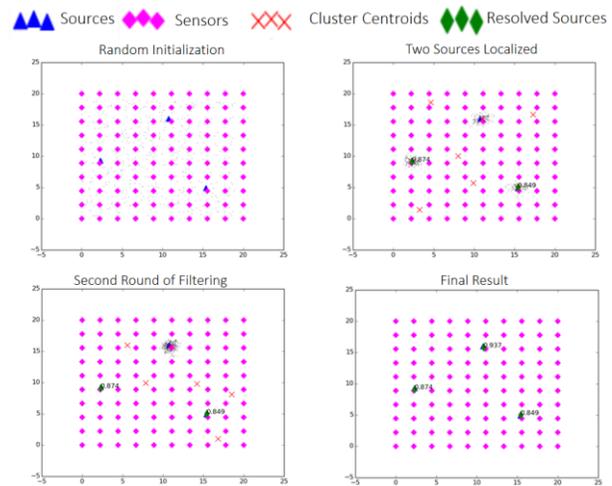

**Figure 6:** Sample run of the algorithm in 2-D space, with 3 sources and 100 readings
(*Algorithm terminates in 6 time steps*)

**Localizing Bulk Sources**

Existing approaches in bulk source localization proceed by approximating a bulk source by its equivalent point source representation [23]. This approximation only holds for very large distances and gives inaccurate and unreliable parameter estimates for short distances, especially in closed environments. Taking inspiration from the theory of electromagnetism, we can claim that particle flux, like any other $1/R^2$ field following the principle of superposition, due to a source distribution can be broken down into its corresponding monopole, dipole, quadrupole moments and so on [24]. The electric potential at a point $p$ due to a charge distribution with total charge $Q_0$, dipole moment $\boldsymbol{P}$ and quadrupole moment $Q_q$, at a distance $r$ from $p$ can be given as:

$$V_0(p) = \frac{1}{4\pi\epsilon_0}\left(\frac{Q_0}{r} + \frac{P \cdot \hat{r}}{r^2} + \frac{Q_q}{r^3}\right) \tag{13}$$

The electric field can be computed as the derivative of this potential as $-\frac{\partial V_0}{\partial r}$. This expression clearly shows why the point source assumptions fail at small values of $r$. Since the proposed algorithm just involves estimation of the parameter vector $\boldsymbol{A}_k$, we can increment the dimension of $\boldsymbol{A}_k$ at the cost of runtime, to achieve similar results. A setup with a wall-type bulk source was simulated in the same environment as before, to validate the idea. By incorporating the dipole contributions alone, the algorithm successfully localizes the distribution with a convincing $c_5$ of 99.2%, in 4 time steps and a runtime of $81s$. The increase in runtime is a result of the extra dimension, which increases the runtime of the mean-shift clustering algorithm.

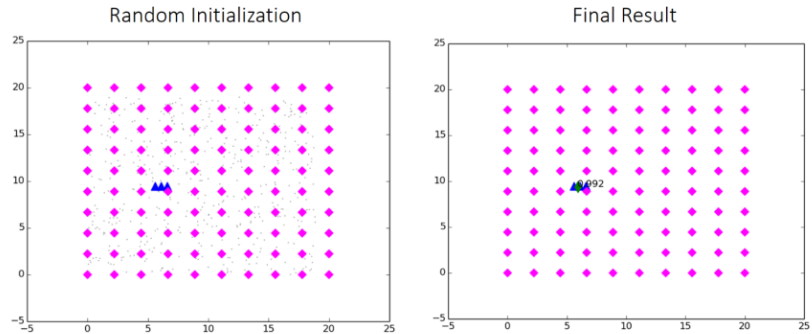

**Figure 7:** Sample run of the algorithm in 2-D space, localizing a wall-type bulk source
*(Algorithm terminates in 3 time steps)*

### Scaling to 3-D

A very important aspect for any localization algorithm is its scalability. In our case, to maintain the same density of coverage in parameter space $\mathcal{A}$, the number of particles required grows exponentially with dimensions. This in turn, would increase the clustering time and overall runtime beyond bounds. To account for these, we propose guided initialization of the particles. Priors from other sensors on the robot, like visual imagery, laser scans etc. can be used to form saliency maps, which can be used to cleverly initialize particles in $\mathcal{A}$. As an example, we consider the case of a closed room from the ASL apartment dataset. Sources were placed at locations corresponding to the sofa and a wooden cabinet, and particles were initialized by under-sampling the point cloud of the scene. The algorithm successfully localizes the two sources (Fig. 8) in 3 time steps, and a runtime of $39s$, hence ascertaining the scalability to higher dimensional regions.

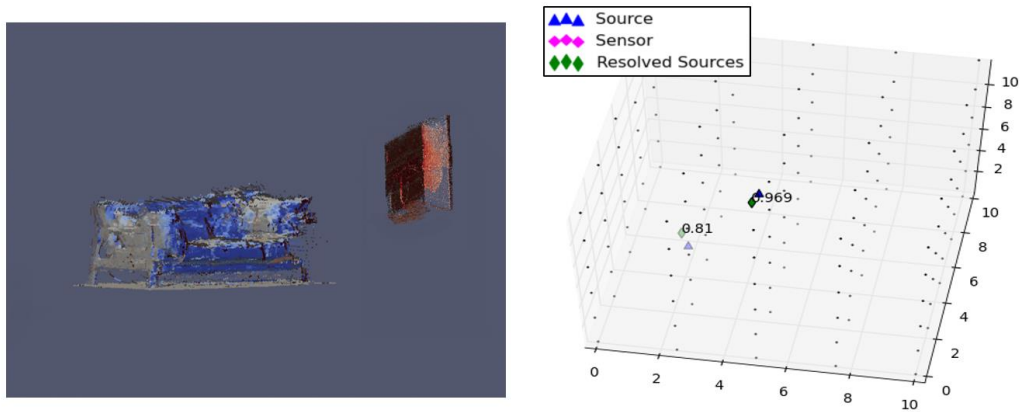

**Figure 8:** Sample run of the algorithm in 3-D space, with guided interpolation using point cloud (left)
*(Algorithm terminates in 3 time steps)*

### Field Test

The algorithm was experimentally verified using the particle flux measurements obtained from the CdZnTe-based Polaris-H camera [14], mounted on top of the *Lilred* ground robot, complete with a Velodyne LIDAR, visual camera, wheel odometry and thermal camera amongst other things. The results of the algorithm can be seen in Fig. 9(d), where it localizes the single point source (100 $\mu Ci$, Cs-137) in 3 time steps (runtime = $37 s$).

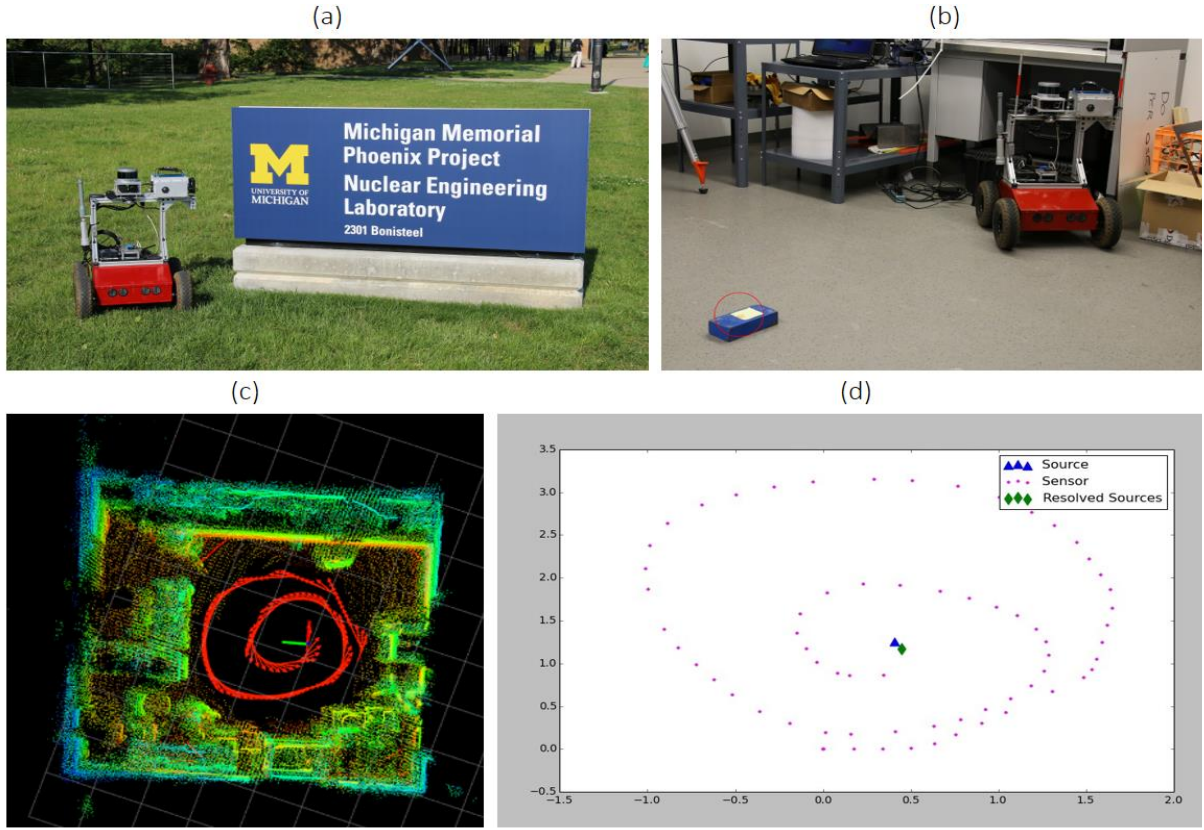

**Figure 9:** Field test at Phoenix Memorial Laboratories: (a) ground robot used; (b) picture from the test (source encircled in red); (c) point cloud of the room (path in red); (d) results of the localization algorithm.

**Quantitative Analysis**

We consider a simulated 2-D environment, with 500 particles and a single point source. The analysis of confidence score $c_3$ and localization error $\epsilon_l$ is given by Fig. 10(a). Here, $\epsilon_l^2 = (\Delta A^{pos})^2 + (\Delta A^{str})^2$ , where $A^{pos}$ is the position in cm, $A^{str}$ is the source intensity in $\mu Ci$, and $\Delta$ refers to the difference between ground truth and estimated source parameters. Fig. 10(a) shows that the drop in localization error is stagnant after 4 time steps, and hence we break every iteration at 3/4 time steps before entering the source estimation stage. Fig. 10(b) shows the 2-D v/s 3-D analysis of the algorithm. With suitable initialization, it is shown that the algorithm can scale seamlessly to higher dimensions, unlike the exponential trend expected otherwise.

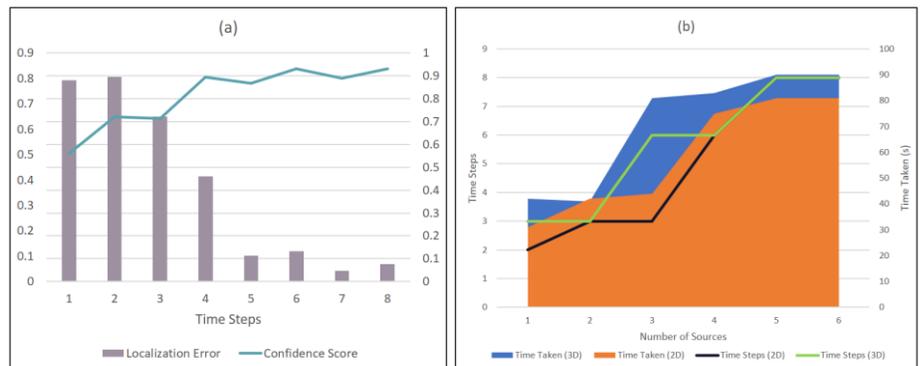

**Figure 10:** (a) Analysis w.r.t. time steps, and (b) 2-D v/s 3-D analysis.

TABLE I: Comparative study of proposed algorithm with state-of-the-art.

| Configuration | Naïve | | | | | | Proposed | | | | | |
|---|---|---|---|---|---|---|---|---|---|---|---|---|
| | Time Steps | Iterations | Localization Error | Precision | Recall | F1 | Time Steps | Iterations | Localization Error | Precision | Recall | F1 |
| 1 Source | 4 | 1 | 0.146 | 0.96 | 1 | 0.980 | 2 | 1 | **0.163** | 0.99 | 0.99 | 0.990 |
| | | | | | | | 3 | 1 | 0.147 | 0.99 | 1 | 0.995 |
| 2 Sources | 4 | 1 | 0.321 | 0.9626 | 0.935 | 0.949 | 3 | 1.11 | 0.289 | 0.975 | 0.975 | 0.975 |
| | | | | | | | 4 | 1.07 | 0.277 | 0.995 | 0.975 | 0.985 |
| 3 Sources | 4 | 1 | 1.259 | 0.759 | 0.7503 | 0.755 | 4 | 1.67 | 0.742 | 0.964 | 0.9663 | 0.965 |
| | 5 | 1 | 1.118 | 0.7695 | 0.8004 | 0.785 | 5 | 1.63 | 0.748 | 0.9766 | 0.973 | 0.975 |
| 4 Sources | 5 | 1 | - | 0.607 | 0.495 | 0.545 | 4 | 2.75 | 1.214 | 0.9629 | 0.9408 | 0.952 |

Table 1 shows a comparative analysis of the proposed algorithm, against the improved particle filter-based implementation suggested by [12]. The environment used is a 2-D square grid of edge length $20m$, with 100 measurements and 1000 particles. *The values mentioned were averaged over 100 executions of the algorithm, for each source configuration.*

We see that the algorithm shows little improvement over the existing method, when the iteration count is 1. This is because the inner loop of the algorithm is fundamentally unchanged. Due to difference in implementation, we do have faster convergence, but the real contributions of the algorithm are clearly visible when the number of iterations go beyond one. Especially for a 3/4 source localization case, as reported, we see a great improvement over the existing algorithm, as the number of iterations approach 2 and more. The improvements in localization error and F1 scores shows the contributions of the outer loop of the algorithm.

As for the running time, the existing approach reports about $0.22s$ per sensor reading, on a 2.40 GHz Intel Core 2 Quad personal computer with 2 GB of RAM, amounting to $43.12s$ for a 196 sensor grid. Running a similar setup with 196 sensor measurements and 2000 particles on a 1.80 GHz AMD A10 Quad processor with 8 GB of RAM, takes $32s$ for 3 time steps, involving just one clustering routine. Even after accounting for the improvement in RAM and lower clock frequency, the proposed algorithm runs faster that the existing algorithm.

## CONCLUSION

We have addressed the problem of robust localization of multiple radiation sources, and distributions thereof, using a mobile robot with consideration of noise and uncertainties. Unlike existing algorithms, the proposed algorithm can also be extended to complex environments, for the localization of bulk sources and is scalable to large environment sizes and higher dimensions. The proposed algorithm is robust to sensor failure, uncertainties in measurement or odometry, and is independent of the path chosen for exploration of the environment. The algorithm can localize very large number of sources with high certainty, and the complexity & running times scale reasonably with the source distribution. Results from simulation and field tests verified the accuracy of our algorithm in multiple realistic scenarios, and show significant improvement over the state-of-the-art.

## ACKNOWLEDGEMENTS

The authors would like to thank Dr. Zhong He and Jiyang Chu at University of Michigan's Orion Radiation Measurement Group for granting access to the Polaris-H radiation detector and for helping us in conducting experiments at their lab. We would also like to thank Michael Lee and Matthew Hanczor for their help with data collection. This work was sponsored by the Department of Energy under award no. **DE-EM0004478.**